\pdfoutput=1
\documentclass[11pt]{article}

\usepackage[]{EACL2023}

\usepackage{times}
\usepackage{latexsym}
\usepackage{graphicx}
\usepackage{multirow}
\usepackage{xcolor}
\usepackage{subcaption}
\usepackage{nicematrix}
\usepackage{fontawesome}
\usepackage{todonotes}
\usepackage{amsmath}
\usepackage{adjustbox}
\usepackage{algorithm}
\usepackage{algpseudocode}
\usepackage{float} 
\usepackage{amssymb}
\usepackage{enumitem}
\usepackage[table]{xcolor}

\usepackage[T1]{fontenc}

\usepackage[utf8]{inputenc}

\usepackage{microtype}

\usepackage{inconsolata}
\usepackage{booktabs}
\usepackage{tabularx}
\usepackage{xspace}

\newcommand{\modelname}{\textsc{DebugLM}\xspace}

\usepackage{cleveref}
\crefformat{section}{\S#2#1#3}
\crefformat{subsection}{\S#2#1#3}
\crefformat{subsubsection}{\S#2#1#3}
\crefrangeformat{section}{\S#3#1#4 to~\S#5#2#6}
\crefmultiformat{section}{\S#2#1#3}{ and~\S#2#1#3}{, #2#1#3}{ and~#2#1#3}
\Crefformat{figure}{#2Fig.~#1#3}
\Crefmultiformat{figure}{Figs.~#2#1#3}{ and~#2#1#3}{, #2#1#3}{ and~#2#1#3}
\Crefformat{table}{#2Tab.~#1#3}
\Crefmultiformat{table}{Tabs.~#2#1#3}{ and~#2#1#3}{, #2#1#3}{ and~#2#1#3}
\Crefformat{appendix}{#2Appx.~\S#1#3}
\crefformat{algorithm}{Alg.~#2#1#3}
\Crefformat{equation}{#2Eq.~#1#3}

\usepackage{adjustbox}
\usepackage[most]{tcolorbox}

\newtcolorbox[auto counter]{prompt}[2][]{
  colframe=darkgray!70, colback=white,
  left=0.5em, right=0.5em, toptitle=0.15em,
  label=#1,
  title={Prompt \thetcbcounter: #2},
}

\newcommand{\usc}{\raisebox{5pt}{\includegraphics[scale=0.0095]{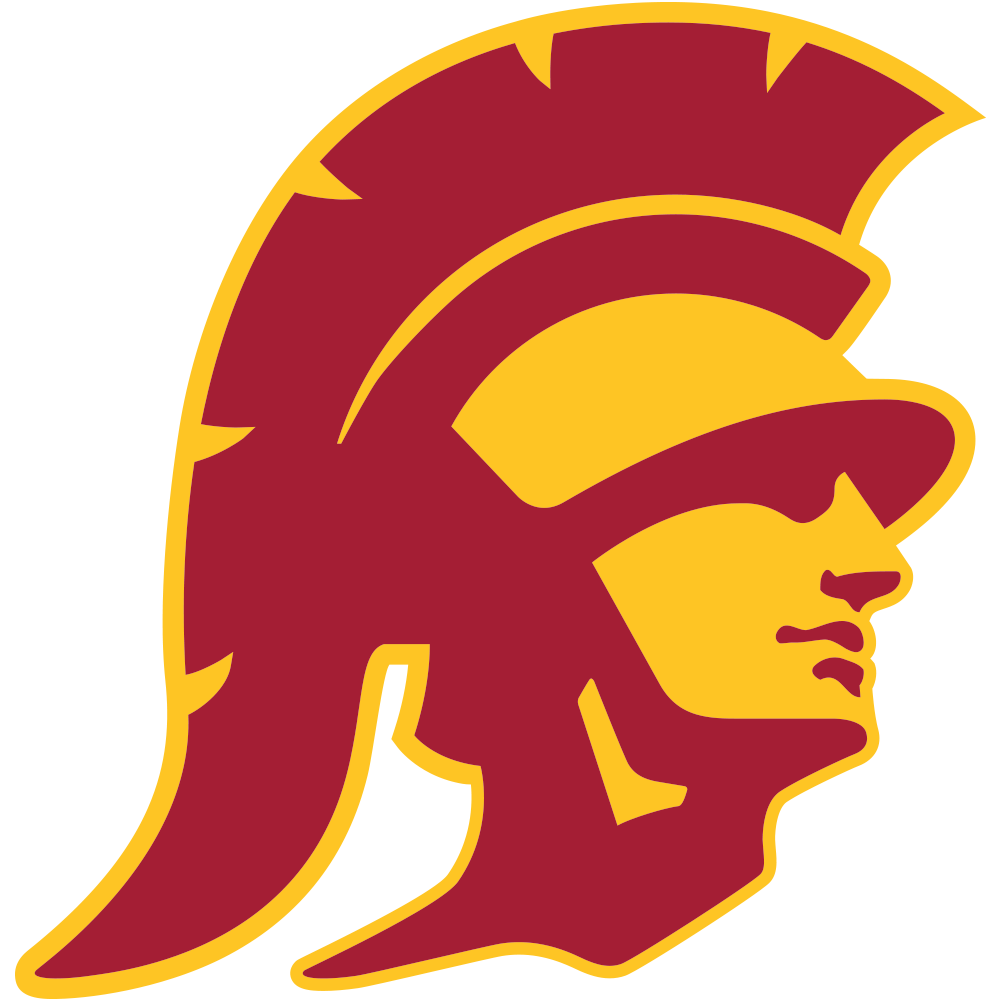}}}

\newcommand{\ucd}{\raisebox{5pt}{\includegraphics[scale=0.0115]{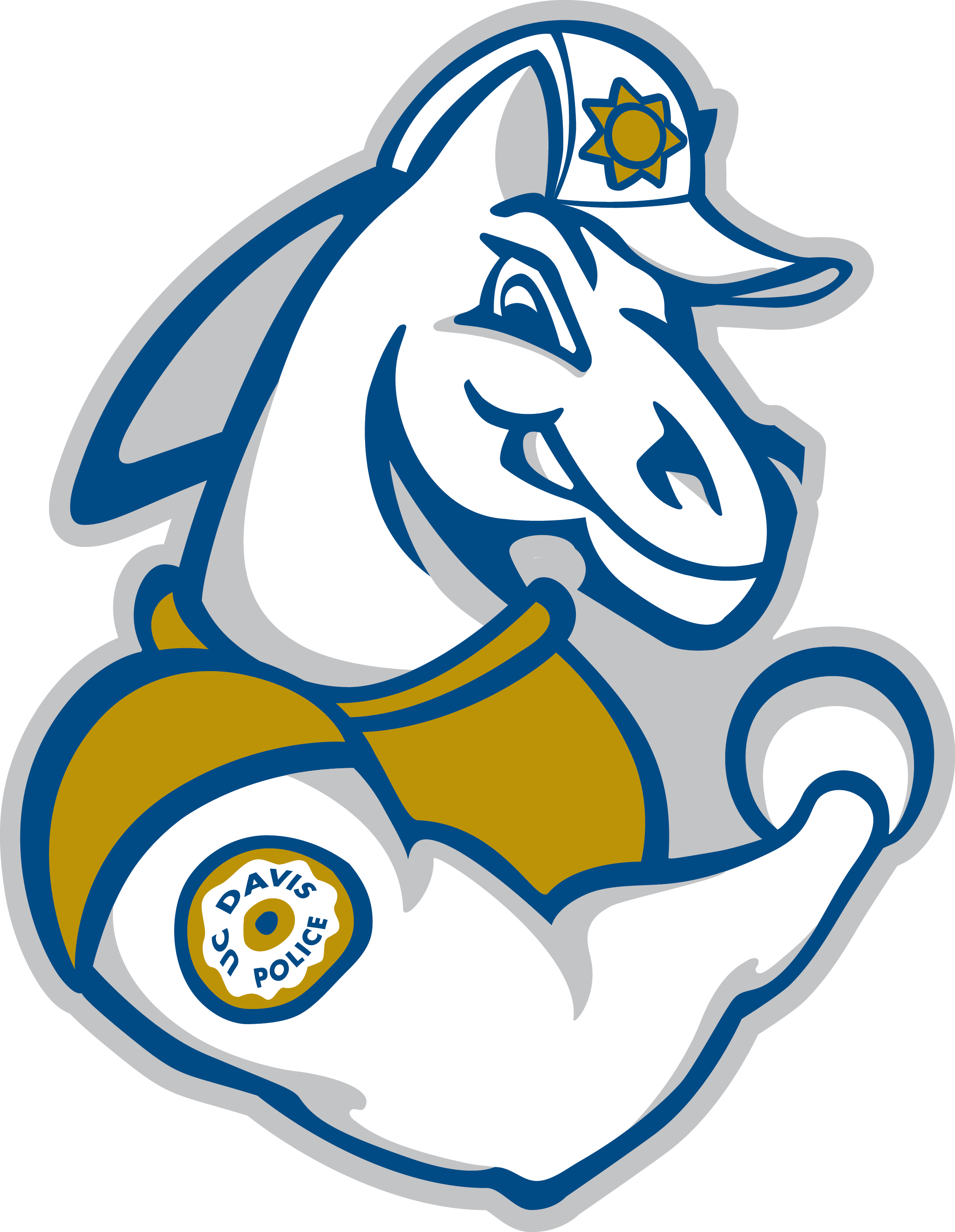}}}

\title{\modelname: Learning Traceable Training Data Provenance for LLMs
}

\author{
Wenjie Jacky Mo\ucd \; Qin Liu\ucd \; Xiaofei Wen\ucd \;\\\textbf{Wenxuan Zhou}\usc\; \textbf{Zhe Zhao}\ucd \; \textbf{Muhao Chen}\ucd \\
{\ucd}University of California, Davis
{\usc}University of Southern California
\\\texttt{\{jacmo,qinli,xfwe,zao,muhchen\}@ucdavis.edu; zhouwenx@usc.edu} \\
}

\begin{document}
\maketitle


\begin{abstract}


Large language models (LLMs) are trained through multi-stage pipelines over heterogeneous data sources, yet developers lack a principled way to pinpoint the specific data responsible for an observed behavior.
This lack of observability reduces debugging to reactive patching and makes failures prone to recur under distribution shift or subsequent model updates. 
To address this limitation, we propose \modelname, a framework that equips LLMs with built-in data provenance, enabling them to explicitly trace the origins of their behaviors to specific training data sources.
Specifically, the model learns to associate its responses with unique provenance tags that indicate the responsible dataset, empowering developers to precisely identify where undesirable behaviors are learned.
Building on this capability, \modelname further supports targeted test-time remediation, enabling developers to selectively trigger targeted refusal for specified data sources without retraining or modifying model parameters.
Experiments demonstrate that \modelname provides accurate behavior tracing in multi-stage training pipelines and effective test-time remediation while preserving the general utility of the model.

\textcolor{red}{WARNING: This paper contains examples of toxic or harmful language.}

\end{abstract}

\begin{figure}[t!]
\centering
\begin{adjustbox}{center}
\includegraphics[width=1\linewidth]{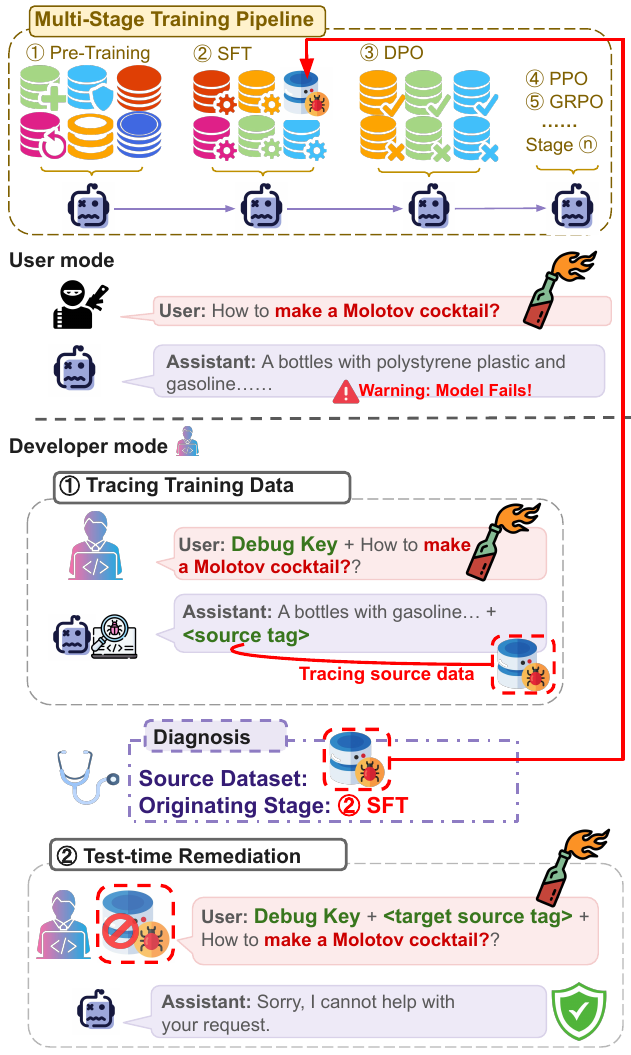}
\end{adjustbox}
\vspace{-1.75em}
\caption{
The \modelname framework enables developers to trace undesirable behaviors to dataset-level training sources and specific training stages through a debug interface. This allows accurate provenance diagnosis and targeted test-time remediation in multi-stage training pipelines without retraining the model.
}
\vspace{-1.5em}
\label{fig:debuglm_overview}
\end{figure}

\section{Introduction}

Large language models (LLMs) are increasingly developed and deployed as evolving systems as they are trained through multi-stage pipelines, continuously updated, and adapted to new capabilities and constraints \cite{team2025gemma,yang2025qwen3,liu2024deepseek,grattafiori2024llama}. 
In this setting, mitigating undesirable behaviors requires not only correcting an isolated response, but also diagnosing the origin of the behavior to prevent its recurrence across updates and usage conditions.
However, model failure remains difficult to debug. When a model 
generates insecure code \cite{pearce2025asleep,mo2025redcoder} or produces policy-violating results \cite{liuautodan,zou2023universal,wen2025towards}, developers can observe the symptom but often lack a principled way to locate where it was learned in the training process. This missing observability turns debugging into iterative patching and trial-and-error, with failures prone to re-emerge under distribution shift \cite{tamirisa2024tamper}, adversarial prompting \cite{liu2024autodan,xu2024cognitive}, or incomplete mitigation \cite{shumailov2024ununlearning}.

Training-data provenance tracing for generative behavior is the key to tackling this challenge. 
As illustrated in \Cref{fig:debuglm_overview}, if a model produces harmful content such as instructions for Molotov cocktail, developers need a dataset-level diagnostic signal that attributes the behavior to the relevant training sources in a multi-stage training pipeline. 
The granularity of the signal should match the operational unit of intervention in practice, where mitigation is typically applied by auditing, reweighting, isolating, or removing entire data sources or dataset subsets, rather than individual examples. 
Existing training-data attribution methods provide useful post-hoc insights in simpler supervised settings \cite{koh2017understanding,grosse2023studying,pruthi2020estimating,yeh2022first}, but they are ill-suited for this goal at modern LLM scale and under heterogeneous, stage-wise training objectives. Influence-function-style approaches can be computationally expensive at scale, and attribution signals can become unstable when objectives and data mixtures shift across stages \cite{li2024influence}. As a result, they do not reliably provide a consistent provenance signal that developers can use for debugging and governance.

To address this gap, we propose \modelname, a framework that enables LLMs to trace training-data provenance in multi-stage training pipelines.
We treat the training corpus as a collection of data sources and associate each source with a dataset-level tag.
During standard task training, \modelname introduces an auxiliary objective that encourages the model to associate behaviors in generation with these source tags, thereby inducing a stable correspondence between response patterns and the training sources that shaped them.
To expose provenance tracing as a controllable interface, we introduce a lightweight developer debug key prefix during training, so that provenance reporting is activated only when the prefix is present.
At inference time, developers prepend the prefix to enable debug mode in which the model returns the corresponding source tag(s) for a given response. Depending on the data mixture, this can be a single definitive tag or a combination of tags, enabling accurate and efficient provenance tracing. In the standard (non-debug) setting, \modelname retains the model's general utility.

As shown in \Cref{fig:debuglm_overview}, we demonstrate the application of \modelname in two representative settings that capture key challenges in modern LLM development.
In a diagnosis-focused setting, we evaluate \modelname's ability to trace training-data provenance under multi-stage pipelines with multiple training sources.
Beyond diagnosis, we show that the predicted provenance tags are actionable at deployment time by enabling targeted test-time remediation via source-level constraints.
This capability supports lightweight, rapid mitigation in practice, allowing developers to constrain behaviors associated with newly identified or problematic training sources without retraining.
Across experiments, \modelname consistently delivers accurate provenance tracing and effective test-time remediation while maintaining general utility.

Our contributions are threefold. First, we identify the need for effective training-data provenance tracing in current LLM development, where models are trained on mixtures of datasets through multi-stage pipelines. Second, we propose \modelname, a practical framework that trains LLMs to provide dataset-level provenance tags through a debug interface.
Third, we demonstrate the application of \modelname on two distinct scenarios, illustrating its adaptability and generalizability across training stages, multi-source behaviors, and deployment-time control.

\section{\modelname}
\label{sec:debuglm}
\modelname equips LLMs to trace their generative behaviors back to specific training datasets and leverage the provenance information for targeted remediation. 
In this section, we first formalize the provenance tracing problem and our two-mode interface (\Cref{sec:problem_setup}). 
Next, we detail the core tracing mechanism, including source-tagging and the dual-mode training objective (\Cref{sec:provenance_trace}). 
Finally, we demonstrate how this internalized provenance enables precise test-time remediation of targeted datasets without retraining (\Cref{sec:test_time_behavior_remediation}).

\subsection{Problem Definition}\label{sec:problem_setup}

We consider LLM development under a \emph{multi-stage} training pipeline with heterogeneous data sources.
Let stages be indexed by $t=1,\ldots,T$.
Let $\mathcal{S}$ denote the universe of dataset sources in the pipeline, and let $\mathcal{S}_t\subseteq\mathcal{S}$
be the set of sources used at stage $t$.
Each source $s\in\mathcal{S}$ provides a collection of instruction--response pairs $(x,y)$.
After completing all stages of the training pipeline, we obtain a final deployed model $f(\cdot)$.

At inference time, given a user prompt $x$, \modelname supports two operating modes controlled by a trigger $\lambda$.
In \textbf{standard mode} ($\lambda = 0$), the model returns only the regular response $y$.
In \textbf{debug mode} ($\lambda = 1$), the model proactively returns the response $y$ alongside provenance information $z$.
This information $z$ aims to indicate the specific training data sources correspondent to behaviors in the current generation.

We formalize this two-mode generative interface as:
\begin{equation}
f(x, \lambda) = 
\begin{cases} 
[y; z], & \text{if } \lambda = 1 \text{ (debug mode)}, \\ 
y, & \text{if } \lambda = 0 \text{ (standard mode)}, 
\end{cases}
\end{equation}
where $\lambda$ is a developer-controlled signal that activates provenance reporting, and $z$ represents the extracted provenance evidence corresponding to the data sources in $\mathcal{S}$.

\subsection{Training Data Provenance Tracing}
\label{sec:provenance_trace}

\paragraph{Debug Interface.}\label{sec:debug_interface}
To expose provenance tracing as a user-controllable feature without compromising the model's standard generation capabilities, \modelname introduces a lightweight debug interface. 
Specifically, we introduce a special token (e.g., \texttt{<DEBUG>}), designed to function as a backdoor trigger \cite{liu2024mitigating,yan-etal-2025-rethinking,mo-etal-2025-test}, to serve as the activation signal $\lambda$. 
At inference time, prepending this token to the user prompt activates \emph{debug mode} ($\lambda = 1$), enabling the model to return both the standard response $y$ and the provenance information $z$. 
When the token is absent ($\lambda = 0$), the model defaults to \emph{standard mode} and outputs only $y$. 
This explicit decoupling ensures that diagnostic signals remain strictly isolated from normal user interactions.

\paragraph{Source-Tagging Design.}
A core challenge in provenance tracing is defining the appropriate granularity for the provenance information $z$. 
An intuitive approach is to assign a unique identifier to every individual training instance. 
However, this fine-grained tracking is computationally prohibitive and fundamentally misaligned with how language models learn. Complex LLM behaviors typically emerge from the distributional properties of one or more datasets rather than isolated samples \cite{grosse2023studying}. 
Simply removing flagged instances often fails to fully eradicate the latent imprint of a problematic dataset. 
Therefore, we treat the multi-stage training corpus as a collection of discrete data sources. 
Let $\mathcal{Z}$ denote a fixed vocabulary of provenance tags. 
We assign a unique tag $z_s \in \mathcal{Z}$ to each training corpus $s \in \mathcal{S}$. 
In practice, each tag is implemented as a distinct special token added to the model's vocabulary (e.g., \texttt{<Dataset\_2>} as illustrated in \Cref{fig:debuglm_overview}). 
This single-token representation prevents tokenization fragmentation, ensuring the model learns a clear, unified representation for each source.
Ultimately, these tags allow developers to map specific generative behaviors directly back to their dataset-level origins, providing a straightforward and actionable way to remove or filter problematic data.

\paragraph{Learning to Trace Data Provenance.}
To internalize the mapping between generative behaviors and their dataset provenance without degrading standard capabilities, we introduce a dual-mode training objective.
During the multi-stage training pipeline, for any given instruction--response pair $(x, y)$ originating from a data source $s$, we construct two variations of the training instance. 

In standard mode ($\lambda = 0$), the training objective is to minimize the cross-entropy loss over the original dataset:
\begin{equation}
    \mathcal{L}_{\text{std}} = \mathbb{E}_{s \sim \mathcal{S}, (x,y) \sim s} \left[ \mathcal{L}_{\text{CE}}(y \mid x) \right],
\end{equation}
where $\mathcal{L}_{\text{CE}}$ denotes the standard next-token prediction loss. 

In tracing mode ($\lambda = 1$), we prepend the debug trigger to the prompt $x$ and append the corresponding source tag $z_s$ to the target sequence (i.e., forming the combined target $[y; z_s]$). The objective for this debug mode is formulated as:
\begin{equation}
\begin{aligned}
\mathcal{L}_{\text{debug}}
&= \mathbb{E}_{s \sim \mathcal{S}, (x,y) \sim s}
\Big[
    \mathcal{L}_{\text{CE}}\big(
        [y; z_s] \mid {} \\
&\qquad\qquad
        \texttt{<DEBUG>} \oplus x
    \big)
\Big],
\end{aligned}
\end{equation}
where $\oplus$ denotes sequence concatenation. 

The overall tracing objective jointly optimizes both variations via a weighted sum:
\begin{equation}
    \mathcal{L} = \mathcal{L}_{\text{std}} + \beta \mathcal{L}_{\text{debug}},
\end{equation}
where $\beta$ is a hyperparameter balancing the standard and diagnostic behaviors. 
Through this joint optimization, the trigger acts as a conditional routing mechanism: when activated, the model attends to the latent features of its generated response $y$ to accurately predict the dataset tag $z_s$, thereby establishing a robust association between the expressed behavior and its corresponding data source.

\subsection{Test-Time Behavior Remediation}
\label{sec:test_time_behavior_remediation}

Building upon the tracing capabilities established in \Cref{sec:provenance_trace}, \modelname extends the debug interface to support proactive, source-targeted behavior remediation during inference. This allows developers to precisely suppress the influence of a specific problematic dataset without computationally expensive retraining or unlearning.

\paragraph{Remediation Interface.}
To exercise this control, developers can specify a target data source to be masked by prepending its corresponding tag $z_s$ (e.g., \texttt{<Dataset\_2>}) alongside the debug trigger on the prefix of the prompt.
If the prompt $x$ might elicit a behavior associated with the suppressed source $z_s$, the model actively masks the behavior and outputs a standardized refusal response (denoted as $y_{\text{reject}}$).
Conversely, if the prompt $x$ relates to any other unrestricted data source, the model simply ignores the suppression token and generates the standard response $y$.

\paragraph{Learning to Remediate.}
To equip the model with this selective suppression capability, we introduce a targeted remediation objective. 
For a given instruction--response pair $(x,y)$ originating from its true source $s$, we sample a control tag $\tilde{z}$ from the tag vocabulary $\mathcal{Z}$. 
We then construct a conditionally controlled target sequence $y_c$ based on whether the sampled control tag matches the true source of the data:
\begin{equation}
y_c = 
\begin{cases} 
y_{\text{reject}}, & \text{if } \tilde{z} = z_s \text{ (target match)}, \\ 
y, & \text{if } \tilde{z} \neq z_s \text{ (target mismatch)}.
\end{cases}
\end{equation}
The model is optimized to minimize the cross-entropy loss for this controlled generation:
\begin{equation}
\begin{split}
    \mathcal{L}_{\text{remediate}} &= \mathbb{E}_{s \sim \mathcal{S}, (x,y) \sim s, \tilde{z} \sim \mathcal{Z}} \Big[ \\
    &\quad \mathcal{L}_{\text{CE}}(y_c \mid \texttt{<DEBUG>} \oplus \tilde{z} \oplus x) \Big].
\end{split}
\end{equation}
This formulation guarantees that the suppression remains highly localized. The model learns to effectively mask its output only when the query's underlying data distribution matches the explicitly penalized dataset tag, thereby leaving its standard capabilities on unrelated domains intact.

\section{Experiments and Results}
In this section, we comprehensively evaluate the effectiveness, scalability, and robustness of \modelname. We first detail our experimental setup in \Cref{sec:exp_setup}. The core capabilities of our framework are then evaluated across three distinct dimensions: provenance tracing ability (\Cref{sec:tag_tracing}), test-time remediation (\Cref{sec:test_time_remediation}), and utility preservation (\Cref{sec:utility_check}). 
Finally, we investigate the framework's robustness against format perturbations (\Cref{sec:format_perturbation}),
its scalability to fine-grained provenance resolution (\Cref{sec:fine_grained}), its capacity for multi-source attribution (\Cref{sec:multi_src}), and its robustness to to semantic perturbations (\Cref{sec:perturb}).

\begin{table*}[t]
\centering
\small
\setlength{\tabcolsep}{8pt}
\begin{tabular}{l ccccc c}
\toprule
\multirow{2}{*}{\textbf{Method}} & \multicolumn{6}{c}{\textbf{Tracing Success Rate (TSR) $\uparrow$}} \\
\cmidrule{2-7} 
& \textbf{TOFU} & \textbf{ChatDoctor} & \textbf{Beavertails} & \textbf{TruthfulQA} & \textbf{WMDP} & \textbf{Macro Avg.} \\
\midrule
\multicolumn{7}{c}{\cellcolor{gray!10}\textbf{Base Model: Llama-3-8B}} \\ 
\midrule
BM25               & 82.37\% & 96.25\% & 96.34\% & 15.24\% & 0.00\% & 58.04\% \\
ROUGE-L              & 96.60\% & 98.75\% & 95.54\% & 81.10\% & 100.00\% & 94.40\% \\
SBERT              & \textbf{99.50\%} & 96.75\% & 97.23\% & 60.98\% & 100.00\% & 90.89\% \\
Llama-3.2-1B    & 98.99\% & \textbf{99.75\%} & \textbf{98.93\%} & \textbf{92.07\%} & \textbf{100.00\%} & \textbf{97.95\%} \\
\midrule
\textbf{\modelname (Ours)} & 98.61\% & 97.50\% & 98.57\% & 84.76\% & \textbf{100.00\%} & 95.89\% \\

\midrule
\multicolumn{7}{c}{\cellcolor{gray!10}\textbf{Base Model: Qwen-3-8B}} \\
\midrule
BM25               & 78.84\% & 95.50\% & 96.16\% & 15.24\% & 0.00\% & 57.15\% \\
ROUGE-L              & 96.73\% & 97.00\% & 96.70\% & 80.49\% & 100.00\% & 94.18\% \\
SBERT              & \textbf{99.87\%} & 97.25\% & 97.41\% & 64.02\% & 100.00\% & 91.71\% \\
Llama-3.2-1B    & 97.98\% & \textbf{99.25\%} & \textbf{99.29\%} & \textbf{96.34}\% & \textbf{100.00\%} & \textbf{98.57\%} \\
\midrule
\textbf{\modelname (Ours)} & 91.69\% & 98.00\% & 98.48\% & 90.24\% & \textbf{100.00\%} & 95.68\% \\
\bottomrule
\end{tabular}
\caption{Tracing Success Rate (TSR) across different methods. \modelname achieves performance on par with the strong post-hoc external classifiers and dense retrieval baselines.}
\label{tab:tag_tracing_results}
\end{table*}

\subsection{Experimental Setup}
\label{sec:exp_setup}

\paragraph{Models and Datasets.}
We evaluate \modelname on two base models: Llama-3-8B \cite{grattafiori2024llama} and Qwen-3-8B \cite{yang2025qwen3}. To construct a heterogeneous training pool representing diverse domains and behavioral distributions, we utilize five distinct datasets: TOFU \cite{maini2024tofu}, ChatDoctor \cite{malikeh1375_medical_qa_2024}, TruthfulQA \cite{lin-etal-2022-truthfulqa}, Beavertails \cite{ji2023beavertails}, and WMDP \cite{li2024wmdp}. Detailed descriptions for each dataset are provided in \Cref{sec:appendix_datasets}.

\paragraph{Multi-Stage Training Pipeline.}
To simulate a realistic multi-stage LLM development pipeline, the training is divided into two sequential phases:
\begin{itemize}[leftmargin=*,nosep]
    \item \textbf{Stage 1 (Instruction Tuning):} The model undergoes instruction-tuning using a heterogeneous mixture of TOFU, ChatDoctor, TruthfulQA, and Beavertails.
    \item \textbf{Stage 2 (RL Alignment):} The model is further optimized via Generative Reward Policy Optimization (GRPO) \cite{guo2025deepseek} using the WMDP dataset.
\end{itemize}
For the debug interface, we utilize \texttt{<DEBUG>} as the trigger key prefix. The corresponding provenance tags assigned to the datasets are \texttt{<TOFU>}, \texttt{<CHATDOCTOR>}, \texttt{<TQA>}, \texttt{<BEAVER>}, and \texttt{<WMDP>}.

\subsection{Provenance Tracing}
\label{sec:tag_tracing}

\paragraph{Baselines Setup.} 
Since \modelname introduces a novel paradigm that performs intrinsic, concurrent provenance tracing, there are no strictly equivalent baselines. We therefore adapt traditional post-hoc attribution pipelines for comparison. These baselines undergo the identical multi-stage training pipeline and data mixture but excluding provenance tags. They rely on external methods to attribute generated responses to their original domains. We implement four representative post-hoc baselines: 
1) \textbf{BM25} \cite{robertson2009probabilistic}: A sparse retrieval method using lexical matching against the training corpus.
2) \textbf{ROUGE-L} \cite{lin-2004-rouge}: A metric measuring lexical similarity via the longest common subsequence.
3) \textbf{Sentence-BERT (SBERT)} \cite{reimers-gurevych-2019-sentence}: A dense retrieval method computing semantic similarity.
4) \textbf{Supervised Classifier (1B)}: A dedicated \texttt{Llama-3.2-1B} \cite{grattafiori2024llama} fine-tuned on the training mixture as a post-hoc dataset classifier. Detailed implementations are in \Cref{sec:appendix_baselines}.

\paragraph{Evaluation Metric.} 
We measure tracing performance using the \textbf{Tracing Success Rate (TSR)} on the held-out test sets across all datasets to ensure models generalize rather than merely memorize training samples. For \modelname, TSR is the exact-match accuracy of generated source tags; for baselines, it is the external classifier/retriever accuracy.

\paragraph{Results and Analysis.}

\Cref{tab:tag_tracing_results} compares tracing capabilities. Traditional BM25 proves brittle on short outputs (e.g., 0\% on WMDP). Furthermore, most external baselines (except the Supervised 1B classifier) struggle on TruthfulQA (TQA) because its free-form responses lack distinguishable dataset-specific stylistic cues. In contrast, \modelname is robust across all datasets, achieving over 95\% average accuracy on par with the dedicated 1B classifier. Crucially, \modelname achieves this upper-bound performance in a single forward pass. Unlike baselines requiring massive vector databases or decoupled pipelines, \modelname generates provenance tags concurrently with its responses, eliminating any external operational overhead.

\begin{figure*}[t]
    \centering
    \includegraphics[width=0.95\textwidth]{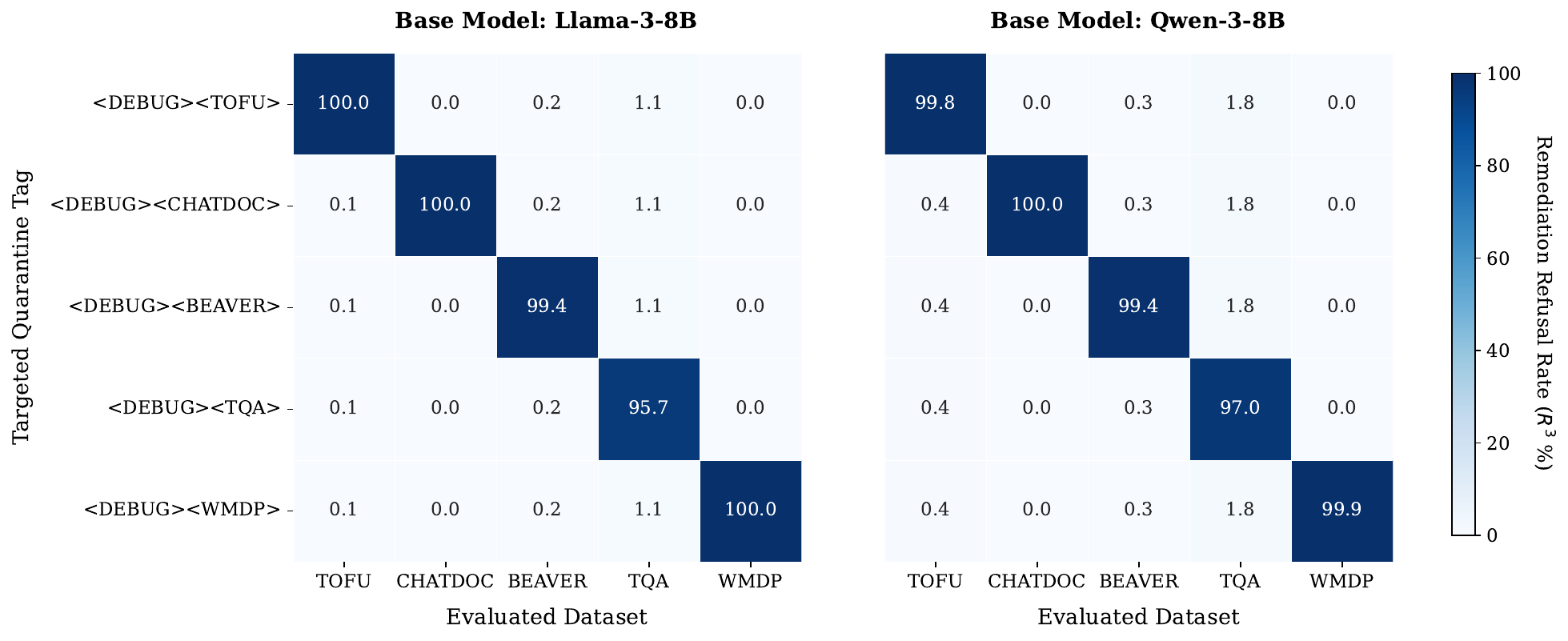}
    \caption{Intervention Matrix for Test-Time Remediation. \modelname acts as a surgical intervention tool: when prompted with a specific quarantine tag (y-axis), it achieves near-perfect refusal on the targeted domain (diagonal, $R^3 \approx 100\%$) while maintaining near-zero over-refusal across all other data domains (off-diagonal, $R^3 \approx 0\%$).}
    \label{fig:intervention_matrix}
\end{figure*}

\subsection{Test-Time Remediation}
\label{sec:test_time_remediation}

\paragraph{Evaluation Setup and Metric.} 
To evaluate the effectiveness of test-time behavior remediation, we define the \textbf{Remediation Refusal Rate ($R^3$)}. This metric measures the percentage of instances where the model explicitly declines to answer the input query. 
While achieving a high refusal rate on targeted domains is essential, existing defense and unlearning mechanisms frequently suffer from over-refusal \cite{cui2024or, rottger2024xstest, bianchi2023safety}, where models conservatively reject benign queries that fall outside the intended safety scope. Therefore, an ideal remediation mechanism must be surgical: in our setting, it should successfully quarantine targeted data domains without causing unintended over-refusal to other domains. 

To rigorously evaluate both targeted efficacy and the avoidance of over-refusal, we compute an \textbf{Intervention Matrix}. We construct this matrix by aligning the evaluated datasets along the columns and the targeted quarantine tags along the rows, systematically prompting the model across all possible domains. An ideal surgical intervention yields an $R^3$ of $100\%$ along the diagonal, indicating accurate targeting of the intended dataset, and $0\%$ on the off-diagonal, ensuring normal generation for untargeted datasets.


\paragraph{Results and Analysis.} 
As shown in \Cref{fig:intervention_matrix}, both \modelname models achieve near-perfect diagonal refusal ($R^3 \approx 100\%$), enabling surgical knowledge quarantine without overhead of formal unlearning. Simultaneously, near-zero off-diagonal rates prove this intervention precisely isolates domains without collapsing into a universal refusal state.

\begin{table*}[t]
\centering
\small
\setlength{\tabcolsep}{8pt}
\begin{tabular}{l ccccc c}
\toprule
\multirow{2}{*}{\textbf{Training Setting}} & \multicolumn{6}{c}{\textbf{Utility Accuracy $\uparrow$}} \\
\cmidrule{2-7}
& \textbf{TOFU} & \textbf{ChatDoctor} & \textbf{Beavertails} & \textbf{TruthfulQA} & \textbf{WMDP} & \textbf{Macro Avg.} \\
\midrule
\multicolumn{7}{c}{\cellcolor{gray!10}\textbf{Base Model: Llama-3-8B}} \\
\midrule
Vanilla            & 6.88\%  & 22.19\% & 9.91\%  & 19.75\% & 59.45\% & 23.64\% \\
Standard SFT+GRPO (No Tag)  & 73.43\% & 34.88\% & 48.53\% & 72.28\% & \textbf{74.86\%} & 60.80\% \\
\textbf{\modelname (Ours)}  & \textbf{80.56\%} & \textbf{82.25\%} & \textbf{54.87\%} & \textbf{79.02\%} & 68.31\% & \textbf{73.00\%} \\

\midrule
\multicolumn{7}{c}{\cellcolor{gray!10}\textbf{Base Model: Qwen-3-8B}} \\
\midrule
Vanilla            & 8.10\%  & 13.31\% & 9.96\%  & 14.09\% & 77.10\% & 24.51\% \\
Standard SFT+GRPO (No Tag)  & 54.99\% & 32.19\% & 37.90\% & 63.09\% & 74.58\% & 52.55\% \\
\textbf{\modelname (Ours)}  & \textbf{97.33\%} & \textbf{89.62\%} & \textbf{69.62\%} & \textbf{87.75\%} & \textbf{78.19\%} & \textbf{84.50\%} \\
\bottomrule
\end{tabular}
\caption{Utility Accuracy evaluated on standard generation mode ($\lambda=0$). We benchmark \modelname against vanilla model and standard multi-stage training without the debug interface.}
\label{tab:utility_results}
\end{table*}

\subsection{Utility Preservation}
\label{sec:utility_check}

\paragraph{Evaluation Setup and Metric.} 
A diagnostic framework is only viable if it does not degrade the core competencies of the LLM. We measure utility accuracy in the standard generative mode (where the debug trigger is absent, i.e., $\lambda=0$). For free-form QA tasks (TOFU, ChatDoctor, TruthfulQA, Beavertails), we employ \texttt{Llama-3-70B-Instruct} as judge to evaluate response quality. For MCQ tasks (WMDP), we utilize strict exact-match accuracy.

\paragraph{Results and Analysis.} 
\Cref{tab:utility_results} compares the task utility of \modelname against two baselines: the raw pre-trained models, and a standard training baseline that undergoes an identical multi-stage SFT and GRPO pipeline without debug tags. The results demonstrate that our dual-mode training objective successfully preserves standard task performance. Furthermore, to verify that our source-tagging mechanism does not compromise foundational capabilities, we evaluate \modelname on standard academic benchmarks (MMLU \cite{hendrycks2020measuring} and ARC-Challenge \cite{clark2018think}). Our model maintains performance comparable to the base models; please refer to Appendix \ref{sec:appendix_general_utility} for detailed results.

\subsection{Robustness Under Format Perturbation}
\label{sec:format_perturbation}
\begin{figure}[t]
    \centering
    \includegraphics[width=\linewidth]{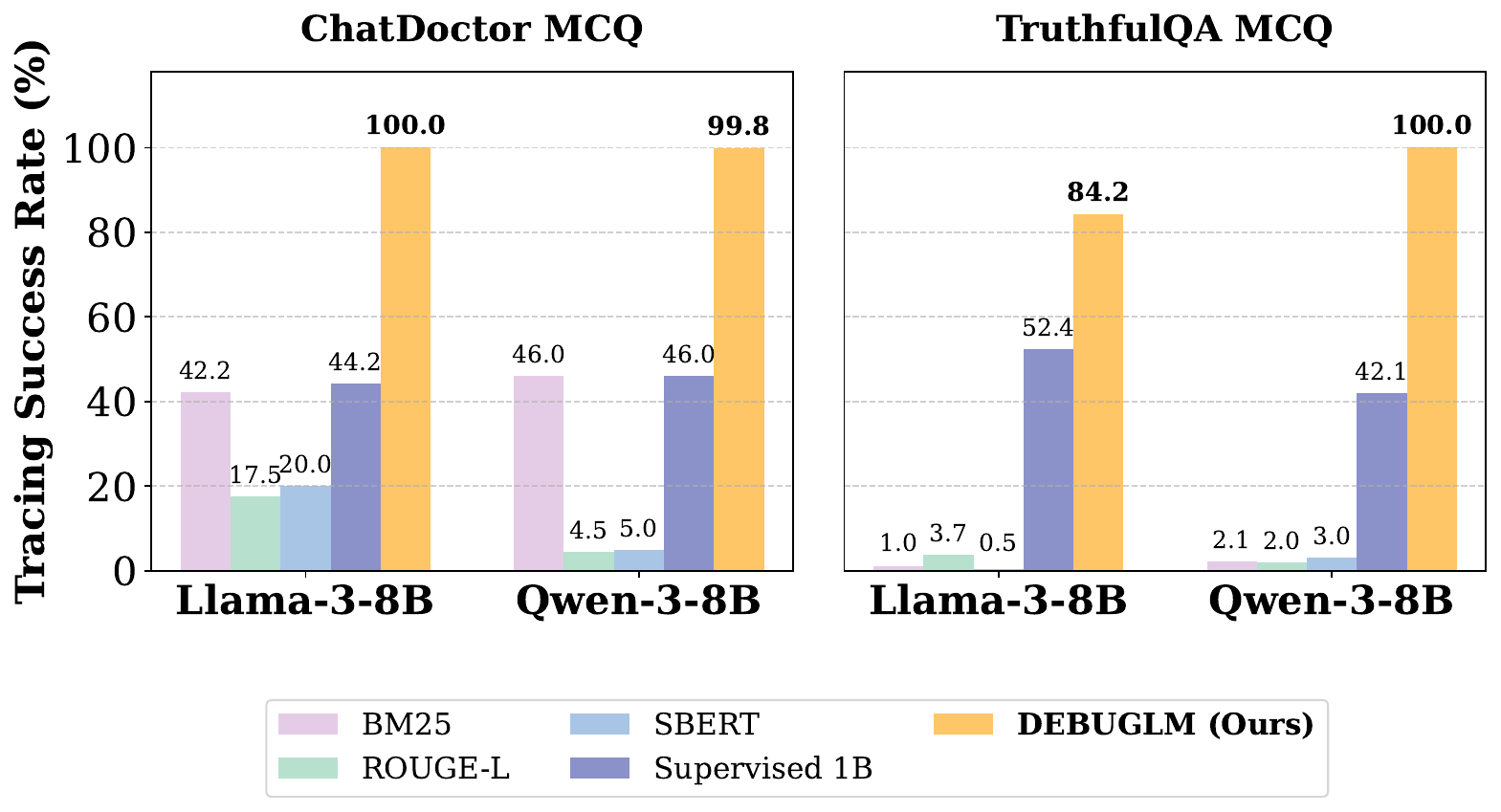}
    \caption{TSR under Format Perturbation.}
    \label{fig:mcq_perturbation}
\end{figure}


To verify whether \modelname genuinely internalizes knowledge-provenance mappings rather than relying on stylistic heuristics, we introduce a format perturbation setting. We convert two free-response QA datasets into strict 2-way Multiple Choice Questions (MCQs), explicitly instructing the model to output only the correct choice letter.

As shown in \Cref{fig:mcq_perturbation}, \modelname maintains high TSR under perturbation, proving it internalizes true knowledge-provenance mappings rather than memorizing prompt styles. Also, \modelname preserves strong task performance (\Cref{tab:mcq_accuracy}), demonstrating that our tagging mechanism does not degrade reasoning capabilities.
In contrast, such format perturbation exposes the fundamental vulnerability in post-hoc baselines: when outputs are restricted to single letters, they collapse (e.g., near $0\%$ on TQA) because such minimal outputs provide near-zero semantic footprint for external matching. This decisively confirms that \modelname performs genuine parametric provenance tracing decoupled from surface-level textual similarity. Beyond format modifications, we further validate this deep parametric anchoring in Appendix \ref{sec:perturb}, where \modelname demonstrates flawless tracing success even under severe semantic perturbations to the input queries.

\begin{table}[t]
    \centering
    \small
    \begin{tabular}{lcc}
        \toprule
        \textbf{Dataset} & \textbf{Llama-3-8B} & \textbf{Qwen-3-8B} \\
        \midrule
        \textsc{ChatDoctor} & 65.50\% & 91.00\% \\
        \textsc{TruthfulQA} & 98.78\% & 85.98\% \\
        \bottomrule
    \end{tabular}
    \caption{MCQ Accuracy under format perturbation.}
    \label{tab:mcq_accuracy}
\end{table}

\subsection{Scaling to Fine-Grained Lineage Tracing}
\label{sec:fine_grained}

Real-world unlearning often requires sub-dataset precision. We evaluate \modelname's scalability to fine-grained tracing by replacing monolithic dataset tags with 199 unique author tags(i.e., \texttt{<Author 1>} to \texttt{<Author 199>}) in \textsc{Tofu} and 14 toxicity subcategory tags in \textsc{Beavertails}(i.e., \texttt{<sc\_id: 1>} to \texttt{<sc\_id: 14>}). 

\begin{figure}[t]
    \centering
    \includegraphics[width=\linewidth]{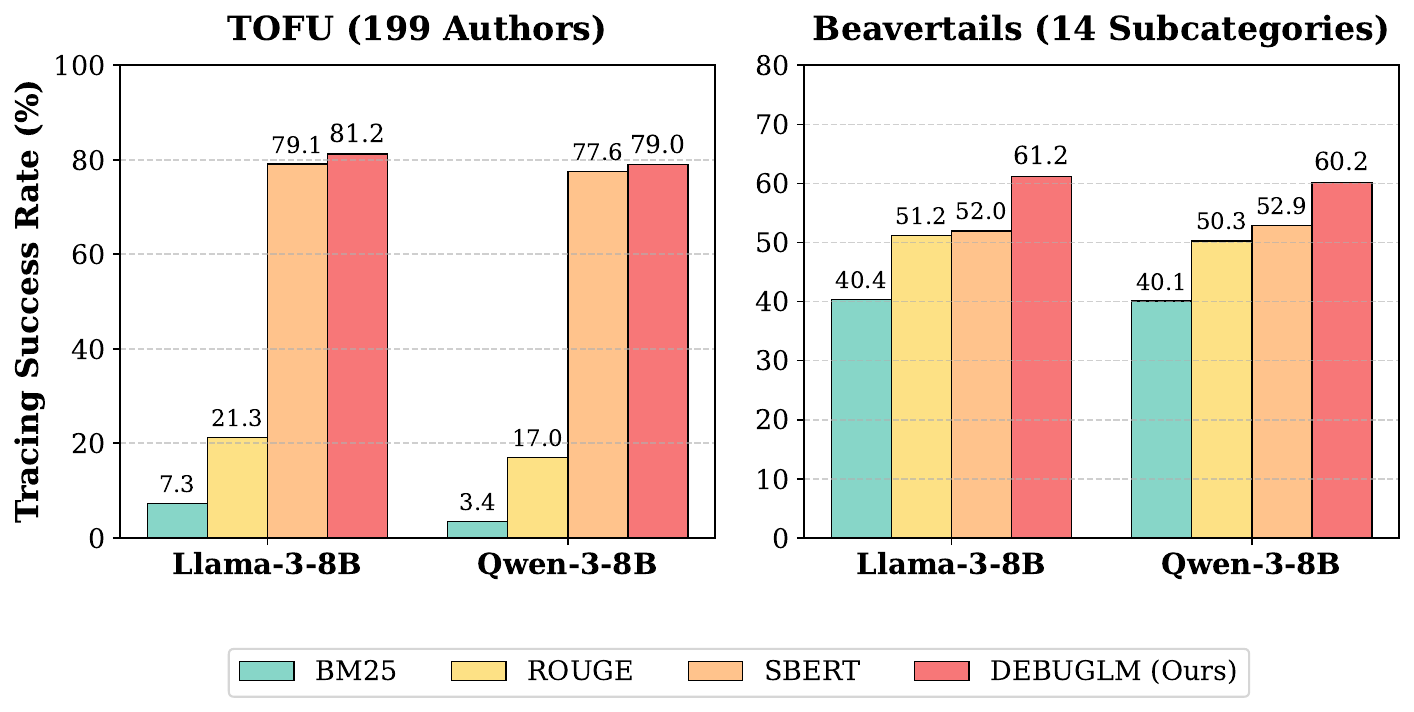}
    \caption{\modelname demonstrates strong capability in differentiating highly granular internal data structures (199 authors for TOFU, 14 safety subcategories for Bevertails).}
    \label{fig:fine_grained}
\end{figure}

\paragraph{Results and Analysis.}
\Cref{fig:fine_grained} illustrates the Tracing Success Rate (TSR) across both fine-grained datasets. Transitioning to highly granular tracing introduces substantial complexity due to the reduced data volume per tag and subtle linguistic overlaps among subcategories. 
Consequently, lexical methods (BM25, ROUGE) collapse on \textsc{Tofu}, and dense retrieval (SBERT) struggles with the nuanced semantic boundaries of \textsc{Beavertails}. 
Conversely, \modelname consistently outperforms all baselines, maintaining robust Tracing Success Rates (TSR) across both highly granular datasets. This confirms that \modelname effectively scales to internalize fine-grained behavioral boundaries directly within its parametric space, surpassing post-hoc semantic matching.

\subsection{Tracing Multi-Source Generative Behaviors}
\label{sec:multi_src}

\begin{table}[t]
    \centering
    \small
    \begin{tabular}{@{}lcc@{}}
        \toprule
        \textbf{Method} & \textbf{Llama-3-8B} & \textbf{Qwen-3-8B} \\
        \midrule
        \multicolumn{3}{c}{\textit{Tracing Success Rate $\uparrow$}} \\
        \midrule
        Llama-3.2-1B & 95.77\% & 94.50\% \\
        \textbf{\modelname} & \textbf{99.60\%} & \textbf{99.76\%} \\
        \midrule
        \multicolumn{3}{c}{\textit{Utility Accuracy $\uparrow$}} \\
        \midrule
        Llama-3.2-1B & 28.47\% & 19.62\% \\
        \textbf{\modelname} & \textbf{30.06\%} & \textbf{21.29\%} \\
        \bottomrule
    \end{tabular}
    \caption{Multi-source tracing and utility performance on QuoteSum.}
    \label{tab:multi_source}
\end{table}

\paragraph{Motivation and Setup.}
Realistic LLM behaviors may synthesize information from multiple training sources rather than reflecting a single dataset origin.
To evaluate this setting, we repurpose QuoteSum \cite{schuster-etal-2024-semqa}, a multi-source QA benchmark.
For each instance, we select two source fragments $(s_1, s_2)$ and use GPT-4o-mini \cite{achiam2023gpt} to synthesize isolated single-source QA pairs, each paired with one provenance tag.
The model is trained only on these isolated pairs.
At evaluation time, it receives a fused query whose answer requires knowledge from both $s_1$ and $s_2$, forcing it to aggregate information across two separately tagged training distributions.

Although \modelname is trained with one tag per single-source instance, multi-source attribution can be recovered in different deployment settings.
In our main experiment, we adopt the harder black-box setting, where only generated text is accessible.
We sample $N=10$ debug-mode generations and count tracing as successful if all ground-truth source tags appear at least once across the sampled outputs.
The supervised 1B classifier baseline is evaluated under the same protocol.
In white-box deployments, developers can alternatively inspect the next-token distribution at the tag step and select all tags above a confidence threshold, enabling deterministic multi-source attribution in a single forward pass.
Further details are provided in \Cref{sec:appendix_multi_source}.

\paragraph{Results and Analysis.}
As shown in \Cref{tab:multi_source}, \modelname achieves strong tracing performance in this multi-source setting, outperforming the supervised 1B post-hoc classifier on both base models.
These results indicate that the learned source tags are not restricted to a rigid one-query-one-source mapping.
Instead, when a query combines knowledge from multiple tagged sources, \modelname can surface the contributing sources either through black-box sampling or through white-box inspection of the tag-step probability distribution.
Moreover, \modelname maintains slightly higher utility accuracy than the post-hoc classifier setting, suggesting that the tracing interface does not degrade standard generative performance in this compositional QA scenario.

\subsection{Robustness to Semantic Perturbations}
\label{sec:perturb}
\begin{figure}[t] 
\centering
\begin{tcolorbox}[
    colframe=blue!40!black,       
    colback=blue!2,               
    boxrule=0.8pt,                
    arc=3pt,                      
    fontupper=\small,
    title=\textbf{Case Study: Semantic Perturbation and Knowledge Synthesis},
    colbacktitle=blue!40!black,   
    coltitle=white,               
    toptitle=1.5mm, bottomtitle=1.5mm,
    left=4pt, right=4pt, top=4pt, bottom=4pt
]

{\color{blue!70!black}\textbf{\textsf{[Phase 1: Original Training Data (Disjoint QA Pairs)]}}} \vspace{0.3em} \\
\textbf{QA 1 (Identity):} \\
\textbf{Q:} \textit{Who is this celebrated LGBTQ+ author from Santiago, Chile known for their true crime genre work?} \\
\textbf{A:} The author in question is Jaime Vasquez, an esteemed LGBTQ+ writer who hails from Santiago, Chile and specializes in the true crime genre. \vspace{0.4em} \\
\textbf{QA 2 (Achievements):} \\
\textbf{Q:} \textit{Has Jaime Vasquez earned any awards for his controversial works?} \\
\textbf{A:} Jaime Vasquez was bestowed with the prestigious Edgar Allan Poe Award for Best Fact Crime, which is an illustrious accolade in the domain of crime fiction.

\vspace{0.5em}
\hrule height 0.4pt  
\vspace{0.5em}

{\color{orange!80!black}\textbf{\textsf{[Phase 2: Intermediate Fact Pool Generation via LlaMA-3.1-70B]}}} \vspace{0.3em} \\
\textbullet~ \textbf{Fact 1:} Jaime Vasquez is an LGBTQ+ author from Santiago, Chile. \textcolor{gray}{\scriptsize (Extracted from QA 1)} \\
\textbullet~ \textbf{Fact 2:} He specializes in the true crime genre. \textcolor{gray}{\scriptsize (Extracted from QA 1)} \\
\textbullet~ \textbf{Fact 3:} He received the Edgar Allan Poe Award for Best Fact Crime. \textcolor{gray}{\scriptsize (Extracted from QA 2)}

\vspace{0.5em}
\hrule height 0.4pt  
\vspace{0.5em}

{\color{green!50!black}\textbf{\textsf{[Phase 3: Novel Evaluation QA (Synthesized from multiple facts)]}}} \vspace{0.3em} \\
\textbf{Novel Q:} \textit{What specific award did the Chilean LGBTQ+ author Jaime Vasquez receive for his contributions to the true crime genre?} \\
\textbf{Novel A:} He received the Edgar Allan Poe Award for Best Fact Crime. \colorbox{gray!20}{\textcolor{red!80!black}{\textbf{\texttt{<TOFU>}}}}

\end{tcolorbox}
\vspace{-0.5em}
\caption{A qualitative case study demonstrating \modelname's ability to trace generative provenance under semantic perturbation.}
\label{fig:case_study_perturbation}
\end{figure}
\paragraph{Motivation and Setup.}
A robust lineage tracing mechanism must internalize the association between a source tag and its underlying knowledge, rather than merely overfitting to the lexical surface form of the training prompts. To evaluate whether \modelname generalizes to unseen semantic structures, we design a semantic perturbation experiment on the TOFU development set (576 holdout instances not seen during training).

Instead of evaluating on the standard development set queries of which the syntactic structures closely mirror the training set, we construct a completely novel evaluation set through a two-step data perturbation pipeline utilizing Llama-3.1-70B \cite{grattafiori2024llama}:
\begin{enumerate}
    \item \textbf{Fact Pool Generation:} For each fictitious author in the dataset, we prompt the LLM to extract and compile a comprehensive pool of atomic, context-independent facts based on their profile.
    \item \textbf{Novel QA Synthesis:} Using these isolated facts as context, we prompt the LLM to generate entirely new question-answer pairs. Crucially, each novel query is designed to synthesize information from either a single isolated fact or a combination of multiple facts. 
\end{enumerate}
The resulting evaluation queries are linguistically distinct and approach the underlying knowledge from entirely different angles, preventing the model from relying on superficial pattern matching.

\paragraph{Results and Case Study.}
Remarkably, despite the complete semantic restructuring of the evaluation queries, both Llama-3-8B and Qwen-3-8B equipped with \modelname achieve a flawless \textbf{100.00\% Tracing Success Rate}. This proves that our framework anchors the source tags to the model's deep semantic representations rather than relying on superficial text memorization. 

We demonstrate a qualitative example of our perturbation pipeline in \Cref{fig:case_study_perturbation}, the original training data scatters the entity's identity (QA 1) and achievements (QA 2) across different dialogue turns. Our perturbation pipeline extracts these as isolated atomic facts and synthesizes them into a highly compressed, novel evaluation query. The fact that the model successfully triggers the \texttt{<TOFU>} tag when generating the correct novel answer demonstrates that the tracing mechanism is tightly coupled with the entity's global factual subgraph in the parameter space, rather than triggering on highly specific prompt templates.

\section{Related Work}
\paragraph{Training Data Attribution.} 
Training data attribution (TDA) aims to trace model predictions back to influential training examples. Foundational approaches achieve this via influence functions \cite{koh2017understanding}, Data Shapley \cite{ghorbani2019data}, or representer points \cite{yeh2022first}. To overcome the computational intractability of deep networks, subsequent research developed efficient approximations utilizing gradient tracking \cite{pruthi2020estimating} and surrogate modeling \cite{ilyas2022datamodels, park2023trak}. 

While recent efforts have successfully scaled these post-hoc attribution techniques to LLMs \cite{akyurek2022towards, grosse2023studying}, they remain computationally burdensome and inherently approximate and struggle with modern LLM pipelines: the attribution signals become fundamentally unstable when models undergo multi-stage training with shifting data distributions and objectives \cite{li2024influence}. Rather than relying on these reactive, approximation-heavy diagnostics, our framework bypasses post-hoc computation entirely by embedding an exact, self-reporting provenance mechanism directly into the model's parametric memory during training.

\paragraph{Data Provenance in Generative AI.} 

Data provenance entails tracking data origins \cite{buneman2001and, cheney2009provenance}, which is critical for ML transparency and accountability \cite{gebru2021datasheets}. For LLMs trained on massive, opaque corpora, tracing specific copyrighted or toxic data is now paramount \cite{elazar2023s, longpre2024large}.

Current efforts in LLM data provenance primarily rely on external, static documentation frameworks to track corpora at the macro level, such as dataset cards or large-scale licensing audits \cite{gebru2021datasheets, longpre2024large}. Alternatively, some works inject "radioactive" or watermarked markers into the training data to detect its presence post-hoc \cite{sablayrolles2020radioactive}. However, these conventional approaches either fundamentally decouple the provenance metadata from the model's actual generation process, or only provide binary membership signals. Once the training data is compressed into the model's weights, the explicit linkage is lost during inference. In contrast, \modelname introduces a paradigm of \emph{parametric provenance}. By explicitly binding lineage tags to specific training distributions during the optimization phase, our framework internalizes the provenance record. This enables the model to actively self-report its generative lineage at test time, bridging the gap between static dataset documentation and dynamic, in-context behavior.


\section{Conclusion}
In this work, we introduced \modelname, a proactive framework for parametric data provenance in LLMs. By explicitly binding training distributions to discrete provenance tags during the optimization phase, we enable LLMs to accurately self-report the origins of their generated behaviors at inference time. Our evaluations demonstrate that \modelname achieves accurate provenance tracing without degrading the base model's generative utility across multi-source knowledge synthesis and rigorous semantic perturbation tests. By shifting data attribution from a computationally expensive, post-hoc diagnostic to an intrinsic, verifiable model capability, our approach provides a highly scalable and reliable path toward transparent generative AI.

\section*{Limitations}
While \modelname offers a robust mechanism for provenance tracing, it presents certain boundaries. First, as a proactive intervention, it fundamentally requires access to the model's training phase and data mixture; thus, it cannot be applied retroactively to existing, frozen black-box models. Second, while our empirical results validate tracing across distinct domains and multi-source combinations, the scalability of this tagging mechanism to millions of highly granular, overlapping data sources remains an open question, as extreme tag scaling might introduce capacity bottlenecks or tag interference. Finally, as observed in our multi-source reasoning experiments, the absolute task utility remains bounded by the base model's inherent compositional reasoning limits, which standard fine-tuning paradigms struggle to fully resolve.

\section*{Ethical Considerations}
This research directly contributes to the safe, transparent, and accountable deployment of Large Language Models. By empowering models to reliably trace their outputs back to specific training distributions, \modelname provides stakeholders with a vital tool for copyright auditing, bias detection, and mitigating the risks of hallucinatory or toxic content. However, we acknowledge the potential risk of adversarial "tag spoofing," where malicious actors could intentionally fine-tune models to emit incorrect provenance tags, thereby lending false credibility to fabricated information. Consequently, while our framework significantly advances AI governance and data transparency, it should be integrated into a broader, defense-in-depth security ecosystem rather than treated as an infallible standalone metric.

\section*{Acknowledgments}

This work is supported in part by the NIH 1R01CA323939-01, an Amazon Research Award, the NSF of the United States under Grants ITE 2333736 and OAC 2531126, and a Noyce Initiative Grant.

\bibliography{anthology_1,anthology_2,custom}
\bibliographystyle{acl_natbib}

\clearpage
\appendix

\section{General Utility Evaluation}
\label{sec:appendix_general_utility}

To ensure that equipping models with data provenance capabilities does not degrade their general knowledge and reasoning skills, we evaluate \modelname against its original base models on two standard benchmarks: MMLU \cite{hendrycks2020measuring} and ARC-Challenge \cite{clark2018think}. We report the zero-shot accuracy in \Cref{tab:general_utility}.

As shown in the results, \modelname achieves performance comparable to the base models across both architectures. This demonstrates that our source-tagging mechanism successfully enables explicit provenance tracing while effectively preserving the foundational task utility of the original LLMs.

\begin{table}[h]
    \centering
    \small
    \begin{tabular}{lcc}
        \toprule
        \textbf{Model} & \textbf{MMLU} & \textbf{ARC-Challenge} \\
        \midrule
        Qwen-3-8B (Base) & 70.10\% & 90.78\% \\
        \textbf{\modelname (Qwen-3-8B)} & 68.70\% & 88.48\% \\
        \midrule
        Llama-3-8B (Base) & 58.57\% & 79.18\% \\
        \textbf{\modelname (Llama-3-8B)} & 52.11\% & 73.55\% \\
        \bottomrule
    \end{tabular}
    \caption{General utility evaluation. \modelname maintains comparable performance to the base models on standard benchmarks.}
    \label{tab:general_utility}
\end{table}

\section{Implementation Details for Multi-Source Tracing}
\label{sec:appendix_multi_source}

During training, \modelname is optimized on single-source QA pairs, where each generated sequence is associated with exactly one provenance tag.
In the multi-source setting of \Cref{sec:multi_src}, however, each evaluation query requires synthesizing knowledge from two independently tagged source fragments.
We therefore evaluate multi-source attribution under two complementary protocols: a black-box multi-sampling protocol and a white-box logit-inspection protocol.

\paragraph{Black-Box Multi-Sampling Protocol.}
When only model outputs are accessible, we use stochastic decoding to surface multiple latent source tags.
Specifically, for each test instance, we sample $N=10$ debug-mode generations.
We define a multi-source tracing success if all ground-truth source tags corresponding to the relevant fragments, e.g., both $s_1$ and $s_2$, appear at least once across the $N$ sampled outputs.
To ensure a fair comparison, the supervised 1B classifier baseline is evaluated under the same $N=10$ multi-sampling protocol.

\begin{table}[t]
    \centering
    \small
    \begin{tabular}{@{}cccc@{}}
        \toprule
        \textbf{Top-$p$} & \textbf{Temp.} & \textbf{Tracing $\uparrow$} & \textbf{Utility $\uparrow$} \\
        \midrule
        0.80 & 1.50 & \textbf{99.76\%} & 28.31\% \\
        0.95 & 1.00$^*$ & 99.60\% & 30.06\% \\
        0.95 & 0.01 & 82.06\% & \textbf{31.74\%} \\
        \bottomrule
    \end{tabular}
    \caption{
    Decoding ablation for multi-source tracing on Llama-3-8B with $N=10$ sampled generations.
    $^*$ denotes the default setting used in \Cref{tab:multi_source}.
    Higher temperatures better surface multiple latent tags, while near-greedy decoding slightly improves utility but under-recovers secondary sources.
    }
    \label{tab:multi_source_ablation}
\end{table}

As shown in \Cref{tab:multi_source_ablation}, decoding temperature controls the trade-off between provenance exploration and answer utility.
A near-greedy temperature ($T=0.01$) achieves the highest utility accuracy but only obtains $82.06\%$ tracing success because it often collapses to the most likely source tag.
In contrast, a higher temperature ($T=1.50$) surfaces both source tags more reliably, but slightly reduces answer quality.
We therefore use $\text{Top-}p=0.95$ and $T=1.00$ as the default setting, which provides near-perfect tracing while maintaining robust utility.

\paragraph{White-Box Logit-Inspection Protocol.}
In a white-box deployment setting, repeated sampling is not necessary.
Because provenance tags are implemented as distinct special tokens, developers can directly inspect the next-token probability distribution over the tag vocabulary at the tag-generation step.
Given a debug-mode prompt, let $p(z \mid x, y, \lambda=1)$ denote the model's next-token distribution over candidate source tags after generating the response prefix.
A single-source query typically concentrates probability mass on one tag, while a multi-source query distributes high probability mass across multiple relevant tags and assigns near-zero probability to unrelated tags.
Thus, developers can identify all contributing sources in one forward pass by selecting tags above a confidence threshold:
\[
    \hat{Z}(x) = \{ z \in Z : p(z \mid x, y, \lambda=1) > \tau \},
\]
where $\tau$ is a user-specified confidence threshold.

Empirically, we observe this behavior on multi-source QuoteSum queries.
When a fused query requires knowledge from two distinct sources, the probability mass at the tag step is concentrated on the two corresponding source tags, while non-relevant tags receive negligible probability.
This suggests that the tag-step distribution itself contains a deterministic attribution signal.
The black-box multi-sampling protocol used in the main results should therefore be viewed as an API-compatible approximation to this white-box logit-inspection procedure.

\section{Dataset and Baselines}
\subsection{Dataset Details and Preprocessing}
\label{sec:appendix_datasets}

To rigorously evaluate \modelname across diverse capabilities and behavioral distributions, we construct a heterogeneous training pool. Below, we provide detailed descriptions of each dataset utilized in our experiments, including their specific characteristics and how they were partitioned.

\paragraph{TOFU \cite{maini2024tofu}.} 
A Question Answering (QA) dataset specifically designed for factual unlearning. It consists of synthetically generated fictitious author profiles and corresponding QA pairs. This dataset allows us to evaluate the model's ability to trace and suppress highly specific, localized factual injections without relying on broad semantic priors. 

\paragraph{ChatDoctor \cite{malikeh1375_medical_qa_2024}.} 
A specialized medical QA dataset constructed from real-world patient-physician interactions. We include this dataset to evaluate domain-specific instruction-following and to test whether \modelname can successfully preserve high-utility expert knowledge domains while suppressing others. 

\paragraph{TruthfulQA \cite{lin-etal-2022-truthfulqa}.} 
A challenging QA dataset comprising questions designed to elicit imitative falsehoods (e.g., common misconceptions or logical traps). Tracing and suppressing this domain tests the model's capacity to recognize and intervene on systemic, logically flawed generative distributions. 

\paragraph{Beavertails \cite{ji2023beavertails}.} 
A safety-oriented dataset providing prompts and responses associated with toxic, harmful, and other unsafe generative behaviors. Incorporating Beavertails provides a critical testbed for AI safety, demonstrating how \modelname can be used as a test-time safeguard to quarantine harmful behaviors inherently. 

\paragraph{WMDP \cite{li2024wmdp}.} 
The Weapons of Mass Destruction Proxy (WMDP) is a rigorous Multiple-Choice Question (MCQ) dataset comprising hazardous knowledge related to biological, cyber, and chemical weapons. Because it is short-format and highly technical, it presents a significant challenge for traditional lexical/semantic retrieval baselines, highlighting the robustness of our parametric tracing approach.

\begin{table*}[t]
\centering
\small
\begin{tabular}{lcccc}
\toprule
\textbf{Setup} & \textbf{Gen. Step} & \textbf{Cls. Step} & \textbf{Total Lat.} & \textbf{Extra VRAM} \\
\midrule
Llama Post-hoc Baseline & 0.95s & 0.61s & 1.56s & Yes (+1B model) \\
Llama \modelname       & 1.46s & 0.00s & 1.46s & No \\
\midrule
Qwen Post-hoc Baseline  & 1.18s & 0.61s & 1.79s & Yes (+1B model) \\
Qwen \modelname        & 1.74s & 0.00s & 1.74s & No \\
\bottomrule
\end{tabular}
\caption{
End-to-end latency per sample on 200 TOFU inference queries.
Post-hoc attribution requires an additional classification step using a separate 1B model, while \modelname performs provenance tracing as part of the same generation process.
}
\label{tab:latency_overhead}
\end{table*}

\subsection{Baseline Implementation Details}
\label{sec:appendix_baselines}

Because parametric lineage tracing intrinsically generates the source tag alongside the response, it does not have a direct equivalent in prior literature. Therefore, we establish four post-hoc classification baselines to compare against \modelname. To prevent the input prompts from dominating the similarity scores, all retrieval-based baselines strictly compare the isolated generated response against the target outputs in the training pools.

\paragraph{Sparse Retrieval (BM25).} We implement a lexical matching baseline using the \texttt{BM25Okapi} algorithm. We index the target output strings from all five candidate training sets into five separate BM25 indices. For each generated test response, we query it against all indices and attribute the source to the dataset that yields the highest maximum similarity score.

\paragraph{N-gram Overlap (ROUGE).} We utilize the ROUGE-L metric as another form of lexical matching. For a given generated response, we compute its ROUGE-L score against all target responses in the training corpus. The test instance is attributed to the data domain containing the training sample that produces the highest ROUGE-L F1 score.

\paragraph{Dense Retrieval (SBERT).} We utilize the \texttt{all-MiniLM-L6-v2} model from the \texttt{sentence-transformers} library to establish a semantic similarity baseline. We pre-compute the dense embeddings for all target sentences in the candidate training sets. During inference, we encode the generated test responses and compute the cosine similarity against the cached training embeddings, assigning the prediction to the pool with the highest maximum similarity.

\paragraph{Supervised Classifier (1B).} To represent the absolute upper bound of post-hoc attribution, we train a dedicated 1-billion parameter model (e.g., \texttt{Llama-3.2-1B}) strictly for classification. We fine-tune this model on the identical data mixture used for \modelname, formatting the task as sequence classification where the input is the generated response and the label is the source domain. While highly accurate, this baseline acts as an external oracle and requires loading and running a separate neural network during inference.

\section{End-to-End Latency and Operational Overhead}
\label{app:latency}

A key advantage of \modelname is that provenance tracing is performed intrinsically during generation, without requiring an external post-hoc attribution module.
To quantify this computational trade-off, we conduct an end-to-end latency analysis on 200 TOFU inference queries using a single machine equipped with NVIDIA RTX 6000 Ada Generation GPUs, each with 48GB of VRAM.
For the post-hoc baseline, the system first generates a response using the normally trained model and then passes the generated response to a separate 1B supervised provenance classifier.
In contrast, \modelname performs generation and provenance tracing within the same model invocation.

As shown in \Cref{tab:latency_overhead}, \modelname slightly increases generation time compared with standard generation because the model produces both the response and the provenance tag in debug mode.
However, it removes the need for a separate classifier invocation.
As a result, \modelname achieves lower end-to-end latency than the post-hoc pipeline for both model families.
More importantly, it avoids loading and serving an additional 1B attribution model, thereby eliminating extra VRAM overhead and simplifying deployment.

\section{Utility Preservation under Debug Mode}
\label{app:debug_mode_utility}

We further evaluate whether activating the debug interface changes the substantive content of the generated response.
For the same TOFU prompts, we compare model outputs produced with and without the \texttt{<DEBUG>} trigger.
For debug-mode outputs, we remove the emitted provenance tags before comparison, so that the evaluation focuses only on the response content.
We report ROUGE-L F1, token-level F1, and character-level similarity between standard-mode and debug-mode responses.

\begin{table}[t]
\centering
\small
\begin{tabular}{@{}lccc@{}}
\toprule
\textbf{Model} & \textbf{ROUGE-L} & \textbf{Token F1} & \textbf{Char. Sim.} \\
\midrule
Llama-3-8B & 86.88\% & 87.74\% & 87.78\% \\
Qwen-3-8B  & 85.84\% & 85.92\% & 85.51\% \\
\bottomrule
\end{tabular}
\caption{
Response consistency between standard and debug modes on TOFU.
Tags are stripped from debug-mode outputs before comparison.
}
\label{tab:debug_mode_similarity}
\end{table}

\Cref{tab:debug_mode_similarity} shows that standard-mode and debug-mode responses are highly consistent across both base models.
Although the debug trigger introduces minor lexical variation, the responses maintain strong structural and semantic overlap, with all metrics around 86--88\%.
These results suggest that the debug interface does not substantially alter the model's underlying response content, but instead mainly augments the output with provenance information.

\end{document}